\documentclass[sigconf,nonacm,pbalance]{acmart}

\AtBeginDocument{%
  }

\usepackage[bottom]{footmisc}
\usepackage{verbatim}
\usepackage{todonotes}
\usepackage{appendix} 
\usepackage{listings} 
\usepackage{xcolor}
\usepackage{adjustbox}
\usepackage{enumerate}
\usepackage{hyperref}

\usepackage{float}
\restylefloat{table}
\usepackage{placeins}
\usepackage{stfloats}
\usepackage{afterpage}
\usepackage{multirow}
\usepackage{fancyvrb}
\usepackage{fvextra}
\usepackage{xcolor}         
\usepackage{todonotes}

\usepackage{fancyhdr}
\fancypagestyle{afterfirst}{
  \fancyhf{}
  \fancyhead[L]{\tiny \textit{HieraRAG}, March 2026, Atlanta, GA, USA}
  \fancyhead[R]{\tiny Chase M. Fensore, Kaustubh Dhole, Jason Fan, Eugene Agichtein, Joyce C. Ho}
  \fancyfoot[C]{\thepage}
}

\begin{document}

\title[Diverse Hierarchical Benchmarking for RAG]{How Fine-Grained Should a RAG Benchmark Be? A Hierarchical Framework for Synthetic Question Generation}

\author{Chase M. Fensore}
\affiliation{%
  \department{Department of Computer Science}
  \institution{Emory University}
  \city{Atlanta}
  \country{USA}}
\email{chase.fensore@emory.edu}

\author{Kaustubh Dhole}
\affiliation{%
  \department{Department of Computer Science}
  \institution{Emory University}
  \city{Atlanta}
  \country{USA}}
\email{kaustubh.dhole@emory.edu}

\author{Jason Fan}
\affiliation{%
  \department{Department of Computer Science}
  \institution{Emory University}
  \city{Atlanta}
  \country{USA}}
\email{jason.fan@emory.edu}

\author{Eugene Agichtein}
\affiliation{%
  \department{Department of Computer Science}
  \institution{Emory University}
  \city{Atlanta}
  \country{USA}}
\email{eugene.agichtein@emory.edu}

\author{Joyce C. Ho}
\affiliation{%
  \department{Department of Computer Science}
  \institution{Emory University}
  \city{Atlanta}
  \country{USA}}
\email{joyce.c.ho@emory.edu}

\renewcommand{\shortauthors}{Chase M. Fensore, Kaustubh Dhole, Jason Fan, Eugene Agichtein, and Joyce C. Ho}

\begin{abstract}
Evaluating retrieval-augmented generation (RAG) systems requires benchmarks that capture diverse question characteristics, yet practitioners lack empirical guidance on which dimensions to vary and at what granularity. We present \textbf{HieraRAG}, a hierarchical framework for studying granularity in RAG benchmark construction, defining optimal granularity as the level that maximizes \emph{discriminative power} (the standard deviation of generation quality across categories) within a given RAG configuration. As a case study, we generate 5,872 synthetic question--answer (QA) pairs from FineWeb-10BT across 3 dimensions (Question Complexity, Answer Type, Linguistic Variation) at 3 granularity levels (2, 4, and 8 categories). With a BM25+Falcon-3-10B pipeline, optimal granularity varies by dimension: complexity benefits from fine-grained distinctions (discriminative power: 0.053) while answer type and linguistic variation peak at medium granularity. We introduce a \textbf{Coherence Ratio} metric to quantify whether fine-grained splits cleanly subdivide parent categories, revealing structural differences across dimensions (Question Complexity: 0.40 vs.\ Answer Type: 1.44). Human evaluation of 110 stratified QA pairs confirms synthetic quality. While these specific findings reflect a single configuration, HieraRAG provides a portable procedure and validation metric for practitioners to determine evaluation granularity within their own RAG settings.
\end{abstract}

\keywords{retrieval-augmented generation, RAG evaluation, synthetic question generation, question answering}


\maketitle

\thispagestyle{plain}
\pagestyle{afterfirst}

\section{Introduction}
\label{sec:intro}

Retrieval-augmented generation (RAG) has become the dominant approach for question-answering (QA) systems, grounding language model outputs in retrieved evidence to improve factual accuracy~\cite{lewis2020retrieval}. As organizations deploy RAG over proprietary corpora, from enterprise documentation to scientific literature, evaluating system performance requires benchmarks that capture diverse question characteristics. Recent work demonstrates that diversity matters for RAG: diverse retrieved content improves answer quality~\cite{wang2025diversity}, and diverse instruction data enhances model capabilities~\cite{liu2025rag}. But as synthetic QA generation becomes more popular for RAG benchmarking ~\cite{datamorgana_filiceGeneratingDiverseQA2025,ip_deepeval_2025}, a fundamental question remains: at what \emph{level of granularity should question characteristics be varied}?

QA evaluation has evolved from factoid extraction~\cite{voorhees_trec-8_2000} toward multi-hop reasoning~\cite{yang_hotpotqa_2018}, diverse answer types~\cite{yona_narrowing_2024}, and varied linguistic formulations~\cite{bolotova_non-factoid_2022}. This reveals natural dimensions along which questions vary: complexity (single- vs. multi-hop reasoning), answer type (factoid vs. abstractive), and linguistic variation (vocabulary alignment, phrasing diversity). Yet existing benchmarks treat these dimensions with inconsistent granularity. HotpotQA categorizes by answer type (Person, Date, Yes/No) but provides limited complexity control~\cite{yang_hotpotqa_2018}. KILT spans task formats but lacks systematic linguistic variation~\cite{petroni_kilt_2021}. Should we distinguish only between ``simple" and ``complex" questions, or use finer categories like ``factoid," ``multi-hop," ``reasoning," and ``comparative"? Finer distinctions may reveal nuanced performance differences, but they increase generation cost and risk  redundant categories. Without empirical guidance, designers risk under-sampling failure modes or diluting evaluation signal with redundant distinctions.

We address this gap through \textbf{HieraRAG}, a hierarchical framework for synthetic RAG QA benchmark construction that systematically varies questions along 3 illustrative dimensions: (1) Question Complexity (QC), (2) Answer Type (AT), and (3) Linguistic Variation (LV). Each dimension is evaluated at 3 granularity levels: coarse (2 categories), medium (4), and fine (8). Unlike prior work that varies answer granularity while keeping questions fixed~\cite{yona_narrowing_2024}, we vary question characteristics themselves to probe how RAG systems respond to realistic query diversity. To validate whether fine-grained splits are well-structured (i.e., cleanly subdividing parent categories rather than introducing unrelated constraints), we introduce a coherence ratio metric analogous to the clustering silhouette coefficient~\cite{rousseeuw1987silhouettes}. We generate 5,872 synthetic questions from the FineWeb-10BT corpus~\cite{penedo2024fineweb}, evaluate them on a \verb|BM25+Falcon-3-10B| pipeline.  To ensure synthetic question quality and validate category assignments, 2 annotators independently evaluate 110 QA pairs for correctness, answerability, and category alignment. Through our experiments, we investigate:
\begin{itemize}
\item \textbf{RQ1:} Which dimensions most differentiate RAG performance?
\item \textbf{RQ2:} Does finer granularity reveal insights or add noise?
\item \textbf{RQ3:} Are fine-grained splits well-structured?
\end{itemize}

Our findings reveal that optimal granularity varies by dimension: the complexity dimension benefits from fine-grained distinctions
(8 categories, \emph{discriminative power}---the standard
deviation of generation quality across categories at a given
granularity level---reaches 0.053) while answer type and
linguistic variation peak at medium granularity (4 categories). The Coherence Ratio reveals structural differences across dimensions (QC: 0.40 vs. AT: 1.44), explaining why some dimensions benefit from finer splits while others plateau. Human evaluation confirms high synthetic quality (98\% acceptable) while low consensus on fine-grained categories (29\% agreement) suggests semantic ambiguity. Preliminary correlation analysis ($r_{pb}=0.24$) indicates that our data-driven Coherence Ratio aligns with this human perception, offering a potential automated proxy for separating categories.

We contribute: (1) a hierarchical framework for determining optimal evaluation granularity in RAG benchmarks, (2) a Coherence Ratio metric for validating hierarchical question category structures, (3) empirical evidence---within a single \verb|BM25+Falcon-3-10B|
configuration---that different question characteristics may
require different granularity levels for diagnostic evaluation, and (4) a dataset of 5,872 hierarchically-organized questions with code for replication.\footnote{\url{https://github.com/fensorechase/rag-diverse-benchmarks-synthetic-qa}}

\section{Methods}
\label{sec:methods}

\begin{figure*}[t]
\centering
\includegraphics[width=0.98\textwidth]{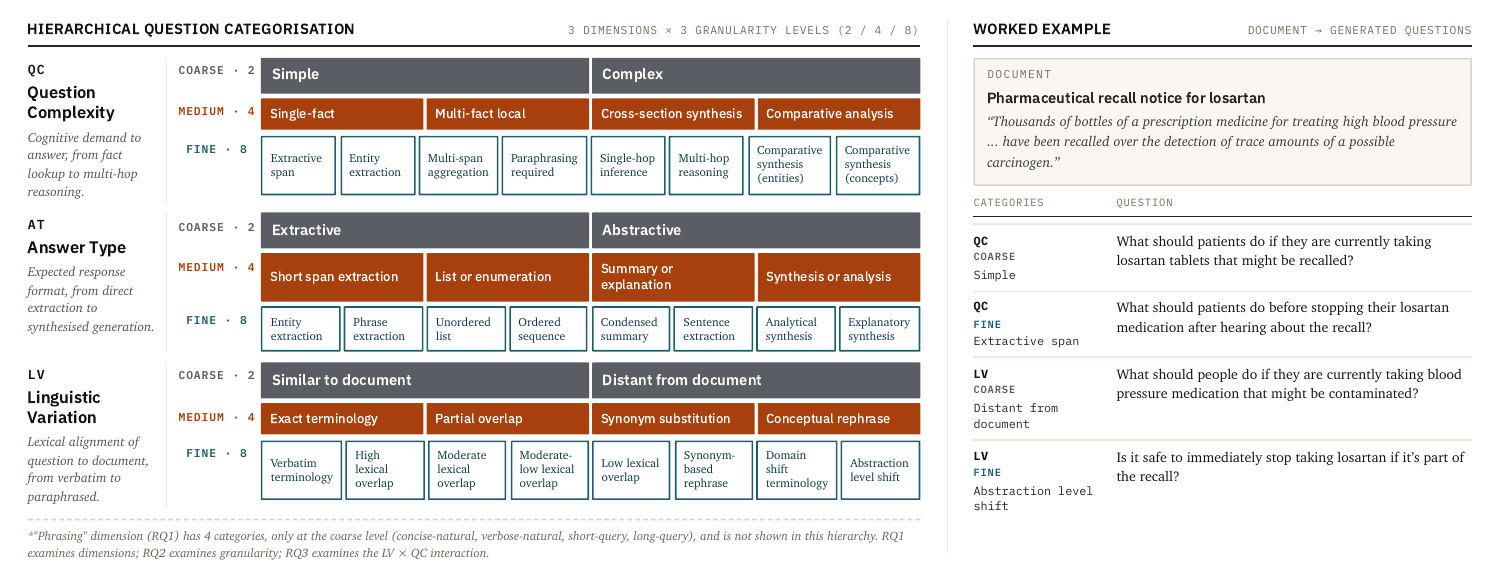}
\vspace{-8pt}
\caption{Hierarchical structure of three question dimensions (QC, AT, LV) across three granularity levels. Each dimension subdivides from coarse (2 categories) to medium (4) to fine (8). User Expertise set to “Novice” for example QAs shown.}
\label{fig:viz_abstract}
\end{figure*}

\subsection{Question Categorization Framework}

To systematically evaluate RAG systems across diverse single-turn question settings with a predefined corpus, we adopted a hierarchical categorization framework capturing 3 key dimensions of question variation (Figure ~\ref{fig:viz_abstract}):
\begin{enumerate}
\item \textbf{Question Complexity (QC)} measures the cognitive demand required to answer questions, ranging from simple fact extraction to multi-hop reasoning.

\item \textbf{Answer Type (AT)} indicates the expected response format, distinguishing between questions requiring direct extraction of information versus those requiring synthesis or generation of new formulations. This dimension builds upon prior work exploring answer type and granularity~\cite{yona_narrowing_2024}.

\item \textbf{Linguistic Variation (LV)} quantifies the lexical alignment between question phrasing and document content, ranging from verbatim terminology to paraphrased concepts requiring semantic understanding.
\end{enumerate}

For each dimension, we define categories at 3 granularity levels: coarse (2 categories), medium (4), and fine (8). Each fine-grained category specializes its parent medium category, which subdivides a coarse category. Categories within each level are mutually exclusive for generation, though our coherence analysis reveals semantic overlap in practice. This structure lets us assess whether finer distinctions reveal additional performance variations, helping practitioners choose appropriate evaluation granularity.

\textbf{Single-dimension assignment as a methodological choice.}
For RQ1 and RQ2, each generated question is assigned to a single dimension during synthesis. We adopt this design to isolate the effect of one dimension at a time, but we acknowledge two consequences. First, real questions can be characterized along all three dimensions simultaneously; the single-dimension assignment is a simplification, not a property of natural questions. Second, when generating along one dimension, the other two dimensions become uncontrolled confounders that can shift retrieval and generation performance independently of the dimension under study, which could compress or amplify the discriminative power reported (e.g., a QC-controlled batch may incidentally contain more distant-vocabulary questions, lowering its mean MAP). RQ3's factorial design (\S\ref{sec:rq3}) probes one such interaction. Post-hoc multi-label tagging via a few-shot classifier is a natural extension.

\subsection{Synthetic Question Generation}

We generated 5,872 questions using DataMorgana~\cite{datamorgana_filiceGeneratingDiverseQA2025}, a tool for creating diverse synthetic QA benchmarks by leveraging Claude 3.5 Sonnet.\footnote{See our supplemental code,~\cite{datamorgana_filiceGeneratingDiverseQA2025}, for details on synthetic QA generation; we note our framework may also leverage other pipelines for synthetic QA generation~\cite{ip_deepeval_2025}.} Questions were generated from documents randomly sampled from FineWeb-10BT~\cite{penedo2024fineweb}, a web corpus with 10B tokens and $\sim$15M documents. Each question includes a DataMorgana-generated reference answer and source document ID to evaluate retrieval and generation quality.

\textbf{User expertise as a held variable.} Across all three research questions, user expertise was randomly specified at generation time (50\% novice, 50\% expert) so that observed performance differences reflect question characteristics rather than user-level variation.
Studying interactions between user persona and question dimensions (QC, AT, LV) is a natural extension for future work.

\textbf{RQ1:} We generated 1,600 questions (400 per dimension) using coarse-level categories from \textbf{4 dimensions}: QC, AT, LV, and Question Phrasing (4 coarse categories only).~\cite{wang2005natural, gupta2015verbose}. 
Each question belongs to one dimension, enabling independent evaluation. Our dimensions were informed by established QA datasets 
~\cite{bolotova_non-factoid_2022, yang_hotpotqa_2018, kwiatkowski_natural_2019, petroni_kilt_2021} and RAG evaluation frameworks~\cite{chen2024benchmarking}. 

\textbf{RQ2:} We generated 3,272 questions across 3 dimensions (QC, AT, LV) at 3 granularity levels (coarse/medium/fine). We focus on these three because they exhibit hierarchical structure suitable for multi-level subdivision. Sample sizes per category vary due to generation stochasticity but provide sufficient power ($n\geq38$ per fine category).

\textbf{RQ3:} We generated 1,000 questions in a 2×2 factorial design, crossing LV (similar/distant) with QC (simple/complex), yielding $\sim$250 questions per cell.

\subsection{RAG System Configuration}
We evaluate a standard two-stage RAG pipeline on an NVIDIA H100. \textbf{Retrieval:} We indexed 512-token chunks of FineWeb-10BT~\cite{penedo2024fineweb} using PyTerrier~\cite{macdonald2021pyterrier} and retrieved $k=10$ documents via \verb|BM25|. \textbf{Generation:} We employed \verb|Falcon-3-10B-Instruct|~\cite{Falcon3} (temp$=0.6$) to generate answers from retrieved contexts, instructing the model to refuse if information was insufficient.

\subsection{Evaluation Metrics}
We evaluate the RAG system along two dimensions: retrieval quality and generation quality~\cite{es2024ragas,chen2024benchmarking}.

\textbf{Retrieval:} Mean Average Precision (MAP) as the primary metric, nDCG@10, and Recall@10 measure whether ground-truth documents appear in top-10 results.

\textbf{Generation:} Cosine similarity (CS) as a primary metric (MiniLM-L6-v2 embeddings)~\cite{reimers2019sentence}, ROUGE-1~\cite{lin2004rouge}, BLEU~\cite{papineni2002bleu} assess answer quality against the reference answer. 

\textbf{Discriminative Power:} For each granularity level, we compute the standard deviation of cosine similarity scores across categories. Higher standard deviation indicates that categories reveal meaningful performance distinctions, justifying finer granularity.

\textbf{Information Content:} We compute normalized mutual information (MI) between category assignments and performance bins. Higher MI indicates category assignments are more informative about system performance~\cite{vinh2010information}. 

\begin{figure}[t]
\centering
\includegraphics[width=\columnwidth]{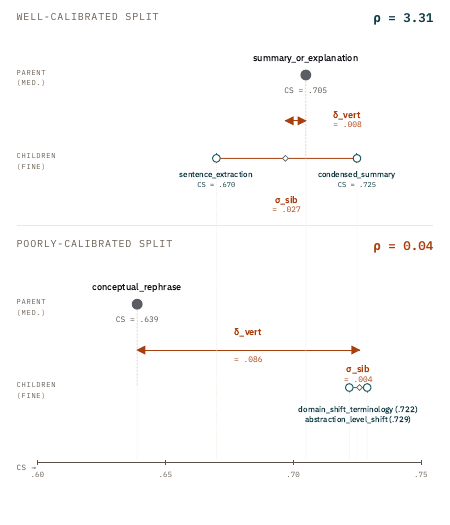}
\caption{Demonstrated Coherence Ratio ($\rho$) calculation for two medium to fine-grained splits within AT and LV dimensions. $\rho > 2.0$ indicates discriminative-yet-aligned children (preferred); $\rho < 1.0$ suggests poor hierarchical structure.}
\label{fig:coherence_example}
\end{figure}

\textbf{Hierarchical Calibration:} Within a dimension, to validate that children categories meaningfully subdivide their parents, we introduce a novel \textbf{\emph{Coherence Ratio}} inspired by the silhouette coefficient in clustering \cite{rousseeuw1987silhouettes}. 
A \emph{split} is one parent-children grouping in the hierarchy (e.g., the medium category \texttt{summary\_or\_explanation} splitting
into the fine categories \texttt{condensed\_summary} and
\texttt{sentence \_extraction}).
Each dimension contains 6 splits in total: 2 coarse-to-medium splits (each coarse parent has 2 medium children) and 4 medium-to-fine splits (each medium parent has 2 fine children). 
For a given corpus and set of questions across nested categories, this metric quantifies whether sibling categories provide non-redundant answer evaluation signals while remaining vertically consistent, and is defined as:
\begin{equation}
\rho_{\text{coherence}} = \frac{\sigma_{\text{sib}}}{\delta_{\text{vert}} + \epsilon},
\end{equation}
where $\sigma_{\text{sib}}$ is the standard deviation of CS across sibling categories (horizontal discrimination), $\delta_{\text{vert}}$ is the absolute difference between parent and mean child CS (vertical consistency).\footnote{$\epsilon=0.001$ prevents division by zero.} High $\rho$ ($\rho > 2.0$) indicates discriminative but aligned siblings (preferred); low $\rho$ ($\rho < 1.0$) suggests poor hierarchical structure (Figure \ref{fig:coherence_example}).


\subsection{Human Validation}
Two annotators 
independently evaluated 110 QA pairs stratified across dimensions and granularity levels. Annotators rated answer correctness (0-3), hallucination severity (0-2), and question answerability (0-3) for all questions. For a subset of 60 questions, annotators additionally validated fine-grained category assignments. 
Inter-annotator agreement for answer correctness was moderate (Fleiss' $\kappa=0.47$). Reference answers proved highly reliable: 98.2\% were acceptable (score $\ge$ 2) and 90.9\% were hallucination-free. 

For category validation, annotators achieved 28.6\% combined agreement (exact or hierarchical parent-match), significantly above the 12.5\% random baseline, but consistent with the inherent difficulty of consistently applying fine-grained definitions manually. Filice et al.~\cite{datamorgana_filiceGeneratingDiverseQA2025} note similar challenges during DataMorgana validation and rely primarily on automated faithfulness filtering rather than fine-grained human category agreement, suggesting that low inter-annotator consensus on fine categories is a known property of synthetic QA benchmarks rather than a flaw specific to our framework.
Practically, low agreement implies that
fine-grained \textit{per-category} performance estimates carry larger effective error bars than coarse-level estimates, and should be interpreted as relative orderings rather than absolute values.

\section{Experimental Results}
\label{sec:results}

\subsection{RQ1: Most Discriminative QA Dimensions}\label{sec:rq1}

To determine which aspects of question variation most strongly differentiate RAG performance, we evaluated 4 dimensions independently using coarse-level categories (Table~\ref{tab:rq2}).


\begin{table}[t]
\centering
\caption{RQ1: Coarse-level dimension comparison. Each dimension uses 2-4 coarse categories. Range $=$ max(CS) $-$ min(CS) across
those categories within a dimension. Note that Range $\neq$ DiscPow in Table \ref{tab:rq1}; CS = Cosine Similarity.}
\label{tab:rq2}
\small
\begin{tabular}{@{}lccccc@{}}
\toprule
\textbf{Dimension} & \textbf{n} & \textbf{MAP} & \textbf{CS} & \textbf{Range} & \textbf{CS Rank} \\
\midrule
LV & 400 & .369 & .695 & \textbf{.077} & 1 \\
Phrasing & 400 & .503 & .703 & .069 & 2 \\ 
AT & 400 & \textbf{.506} & \textbf{.711} & .059 & 3 \\
QC & 400 & .478 & .704 & .010 & 4 \\
\bottomrule
\end{tabular}
\end{table}

LV emerged as the most discriminative dimension (range=0.077) despite achieving the worst absolute performance (MAP=0.369, CS=0.695), while AT achieved the best performance (MAP=0.506, CS=0.711) yet showed lower discrimination (range=0.059).

\subsection{RQ2: Impact of Categorization Granularity}\label{sec:rq2}
We evaluated whether increasing categorical granularity reveals additional performance distinctions (Table~\ref{tab:rq1}). Discriminative power (DiscPow) is measured as the standard deviation of cosine similarity scores across categories within a granularity level; higher DiscPow indicates that member categories provide greater diagnostic resolution for identifying system capabilities and limitations.

\begin{table}[t]
\centering
\caption{RQ2: Discriminative power and performance over granularity levels. QC=Question Complexity; AT=Answer Type; LV=Linguistic Variation; C=Coarse (2 cat.); M=Medium (4 cat.); F=Fine (8 cat.). DiscPow=$\sigma$(CS) across categories; NMI=norm. mutual info. $\Delta$: change from parent level.}
\label{tab:rq1}
\small
\setlength{\tabcolsep}{4pt}
\begin{tabular}{@{}llcccc|cc@{}}
\toprule
\textbf{Dimension} & \textbf{Gran.} & \textbf{n} & \textbf{MAP} & \textbf{CS} &  \textbf{NMI} &\textbf{DiscPow} & \textbf{$\Delta$} \\
\midrule
\multirow{3}{*}{QC} & C (2) & 333 & .518 & \textbf{.712} &  .008 &.007 & -- \\
& M (4) & 333 & \textbf{.564} & .709 &  .027 &.035 & \textbf{4.8×} \\
& F (8) & 446 & .562 & .708 &  \textbf{.040} &\textbf{.053} & \textbf{1.5×} \\
\midrule
\multirow{3}{*}{AT} & C (2) & 333 & .491 & .692 &  .028 &.024 & -- \\
& M (4) & 333 & \textbf{.514} & \textbf{.699} &  .026&\textbf{.044} & \textbf{1.8×} \\
& F (8) & 382 & .504 & .684 &  \textbf{.035} &.037 & -16\% \\ 
\midrule
\multirow{3}{*}{LV} & C (2) & 333 & .412 & .672 &  .049 &.039 & -- \\
& M (4) & 333 & .343 & .671 & \textbf{.054} &\textbf{.047} & \textbf{1.2×} \\
& F (8) & 446 & \textbf{.446} & \textbf{.699} &  .020 &.031 & -33\% \\ 
\bottomrule
\end{tabular}
\end{table}


\textbf{QC} showed monotonic improvement across all three levels (DiscPow from 0.007 at coarse to 0.053 at fine), with a 4.8$\times$ gain from coarse to medium and an additional 1.5$\times$ gain at fine granularity. This indicates that fine-grained complexity distinctions (e.g., single-hop inference vs. multi-hop reasoning) capture meaningful performance variation beyond coarse simple/complex categorization.
In contrast, \textbf{AT} and \textbf{LV} peaked at medium granularity (0.044 and 0.047 respectively) and declined at fine (0.037 and 0.031). This suggests \textbf{diminishing returns beyond 4 categories} for these dimensions. Here, finer distinctions add noise rather than signal.
The normalized MI analysis confirms these patterns: QC increases monotonically but LV peaks at medium.

\textbf{Hierarchical calibration analysis} of the 6 splits per dimension reveals that fine-grained splits are not uniformly well-structured across dimensions (mean $\rho_{\text{coherence}}$=0.73$\pm$0.89). AT shows the best-calibrated split for \textit{summary\_or\_explanation} (coherence=3.31), where children (\textit{condensed summary}, \textit{sentence extraction}) discriminate strongly while aligning with their parent. Conversely, QC shows poor calibration (mean coherence=0.40) despite achieving the highest discriminative power at fine granularity (DiscPow=0.053).
While the 8 fine QC categories successfully partition performance space, they do not cleanly subdivide their medium-level parents.

This also explains why AT's discriminative power \textit{decreases} at fine granularity (from 0.044 to 0.037). Though there is one high-quality split (\textit{summary\_or\_explanation}), the other three AT medium-to-fine splits have low mean coherence (0.82), creating fine categories that cluster together rather than expanding performance coverage. LV's \textit{conceptual\_rephrase} split exhibits the poorest calibration, with high vertical deviation ($\delta_{\text{vert}}$=0.086) indicating fine children (domain shift terminology, abstraction level shift) stray substantially from their parent's performance profile. These poorly calibrated splits introduce constraints beyond simple subdivision, consistent with hierarchy not being strictly taxonomic. Validating $\rho_{\text{coherence}}$, we observed a positive correlation ($r_{pb}=0.24, p=0.21$) between a split's Coherence Ratio and human agreement, suggesting that $\rho_{\text{coherence}}$ is a promising proxy for semantic separation.

\subsubsection{Which Categories Are Hardest?}

Within fine granularity, we observe substantial performance variation (Table~\ref{tab:rq1_fine}). For example in QC, extractive span questions (CS=0.619) underperform comparative synthesis (concepts) (CS=0.766) by 24\%. 


\begin{table}[t]
\centering
\caption{RQ2 Fine-grained: Top-2 and bottom-2 performing categories per dimension by generation quality (CS).}
\label{tab:rq1_fine}
\small
\begin{tabular}{@{}llccc@{}}
\toprule
\textbf{Dimension} & \textbf{Category} & \textbf{n} & \textbf{MAP} & \textbf{CS} \\
\midrule
\multirow{5}{*}{QC}
& comparative\_synthesis\_concepts & 50 & \textbf{.566} & \textbf{.766} \\
& single\_hop\_inference & 59 & .495 & \textbf{.755} \\
\cmidrule(lr){2-5}
& entity\_extraction & 62 & \textbf{.606} & .636 \\
& extractive\_span & 55 & .523 & .619 \\
\midrule
\multirow{5}{*}{AT}
& explanatory\_synthesis & 48 & .395 & \textbf{.738} \\
& condensed\_summary & 44 & .349 & \textbf{.725} \\
\cmidrule(lr){2-5}
& phrase\_extraction & 46 & \textbf{.513} & .647 \\
& ordered\_sequence & 38 & \textbf{.547} & .625 \\
\midrule
\multirow{5}{*}{LV}
& abstraction\_level\_shift & 57 & \textbf{.451} & \textbf{.729} \\
& domain\_shift\_terminology & 60 & \textbf{.333} & \textbf{.722} \\
\cmidrule(lr){2-5}
& synonym\_based\_rephrase & 59 & .293 & .683 \\
& low\_lexical\_overlap & 61 & .225 & .624 \\
\bottomrule
\end{tabular}
\end{table}

Notably, several "complex" children yield higher RAG CS than "simple" children (e.g., comparative\_synthesis\_concepts CS=0.766 vs. extractive\_span CS=0.619), and distant-vocabulary categories achieve high generation scores (CS=0.722–0.729) despite poor retrieval (MAP=0.225–0.333).

\subsection{RQ3: Interaction Between LV and QC}\label{sec:rq3}

To assess whether vocabulary mismatch amplifies complexity effects, we conducted a 2×2 factorial experiment on LV×QC (Table~\ref{tab:rq3}).

\begin{table}[t]
\centering
\caption{RQ3: LV×QC interaction. Gap shows difference (simple to complex) within LV level.}
\label{tab:rq3}
\small
\begin{tabular}{@{}llcc|cc@{}}
\toprule
\textbf{LV} & \textbf{QC} & \textbf{n} & \textbf{MAP} & \textbf{CS} & \textbf{Gap} \\
\midrule
Similar & Simple & 252 & .547 & .637 & -- \\
Similar & Complex & 265 & .566 & .672 & +.035 \\
\midrule
Distant & Simple & 265 & .237 & .578 & -- \\
Distant & Complex & 218 & .164 & .578 & +.000 \\
\midrule
\multicolumn{6}{l}{\textit{Main effects:}} \\
Similar (pooled) & -- & 517 & .557 & .655 & -- \\
Distant (pooled) & -- & 483 & .201 & .578 & \textbf{--.077} \\
\bottomrule
\end{tabular}
\end{table}
LV shows a strong main effect ($\Delta$=0.076, 13\% relative), while complexity shows a weak effect ($\Delta$=0.018, 3\%). The vocabulary gap is consistent across complexity levels: similar vocabulary helps complex questions by +0.035 CS, while distant vocabulary shows no complexity effect (+0.0). This suggests \textbf{additive rather than interactive effects}. Here, LV and QC are largely independent.

The interaction pattern reveals that when vocabulary is mismatched, complexity becomes irrelevant because retrieval has already failed (MAP=0.164–0.237 for distant vocabulary vs. 0.547–0.566 for similar). Complexity distinctions only matter when the system successfully retrieves relevant documents.

\section{Discussion and Conclusion} \label{sec:discussion}
Our hierarchical framework reveals that optimal evaluation granularity varies by dimension. QC benefits from fine-grained distinctions (8 categories) while AT and LV peak at medium (4 categories). This challenges one-size-fits-all  benchmark design and demonstrates how single-level evaluation can obscure critical patterns, as seen where LV dominated at coarse granularity (RQ1), yet QC showed strongest discriminative power hierarchically (RQ2). The additive effects of vocabulary and complexity (RQ3) also support independent dimension design.

\textbf{Calibration for Validation.} We introduce the Coherence Ratio to validate whether fine splits are well-structured. 
High coherence (e.g., AT's \textit{summary\_or\_explanation}: 3.31) indicates discriminative yet aligned children, whereas low coherence (e.g., QC mean: 0.40) reveals redundant distinctions. The correlation between Coherence Ratio and human agreement ($r_{pb}=0.24$) suggests that our metric captures true semantic split boundaries. Practitioners can use $\rho<1.0$ as a signal to refine category definitions or reduce granularity before large-scale QA benchmark generation.

\textbf{Generalization and Limitations.}  Several limitations
should be acknowledged during interpretation of our results: first, all experiments use a single retriever (\verb|BM25|), generator (\verb|Falcon-3-10B|), and corpus (FineWeb-10BT). The specific granularity findings should be read as illustrative for this configuration rather than as general properties of RAG benchmarking; \verb|BM25|'s lexical sensitivity likely amplifies the LV signal at coarse level (Table~\ref{tab:rq1}), and a dense or hybrid retriever may shift which dimensions are most discriminative. The portable contribution is HieraRAG's procedure (hierarchical design, discriminative power, Coherence Ratio), not the category-level outcomes. 
Second, single-dimension assignment in RQ1/RQ2 leaves the other two dimensions as uncontrolled confounders that can inflate or deflate estimated discriminative power; this could be addressed using full-factorial coverage with post-hoc multi-label tagging via few-shot LLM classifiers.
Third, correlation of Coherence Ratio with human agreement is positive but non-significant ($r_{pb}{=}0.24$, $p{=}0.21$, $n{=}60$); we treat it as a useful diagnostic that requires further validation. 
Finally, because properties of synthetic queries differ from human ones,~\cite{zendel2025comparative}, future work should evaluate granularity findings against human-written benchmarks.

\begin{acks}
This work is supported by the National Science Foundation (NSF) grant IIS-2145411 and CISE Graduate Fellowships under Grant No. 2313998. Any opinions, findings, and conclusions or recommendations expressed in this material are those of the author(s) and do not necessarily reflect the views of the National Science Foundation.
\end{acks}



\bibliographystyle{ACM-Reference-Format}
\bibliography{Z-sample-base}


\end{document}